\documentclass{article}

\usepackage[final, nonatbib]{nips_tmp}

\usepackage[utf8]{inputenc} % allow utf-8 input
\usepackage[T1]{fontenc}    % use 8-bit T1 fonts
\usepackage{hyperref}       % hyperlinks
\usepackage{url}            % simple URL typesetting
\usepackage{booktabs}       % professional-quality tables
\usepackage{amsfonts}       % blackboard math symbols
\usepackage{nicefrac}       % compact symbols for 1/2, etc.
\usepackage{microtype}      % microtypography
\usepackage{graphicx}
\usepackage{amsmath}
\usepackage{amssymb}
\usepackage{bm}

\newcommand{\mname}[1]{\mbox{\textbf{#1}}}

\title{Sparse Coding on Stereo Video for Object Detection}

\author{
    Sheng Y. Lundquist\\
    Portland State University\\
    New Mexico Consortium\\
    \texttt{shenglundquist@gmail.com}
    \And
    Melanie Mitchell\\
    Portland State University\\
    Santa Fe Institute\\
    \And
    Garrett T. Kenyon\\
    Los Alamos National Laboratory\\
    New Mexico Consortium
}

\begin{document}

\maketitle              % typeset the title of the contribution

\begin{abstract}
Deep Convolutional Neural Networks (DCNN) require millions of labeled training examples for image classification and object detection tasks, which restrict these models to domains where such datasets are available. In this paper, we explore the use of unsupervised sparse coding applied to stereo-video data to help alleviate the need for large amounts of labeled data. We show that replacing a typical supervised convolutional layer with an unsupervised sparse-coding layer within a DCNN allows for better performance on a car detection task when only a limited number of labeled training examples is available. Furthermore, the network that incorporates sparse coding allows for more consistent performance over varying initializations and ordering of training examples when compared to a fully supervised DCNN. Finally, we compare activations between the unsupervised sparse-coding layer and the supervised convolutional layer, and show that the sparse representation exhibits an encoding that is depth selective, whereas encodings from the convolutional layer do not exhibit such selectivity. These result indicates promise for using unsupervised sparse-coding approaches in real-world computer vision tasks in domains with limited labeled training data.
\end{abstract}

\section{Introduction} \label{sec:introduction}
Over the last decade, Deep Convolutional Neural Networks (DCNN) trained with supervised learning have emerged as the dominant paradigm for computer vision. These networks have shown impressive results on computer vision tasks such as image labeling and object detection (e.g.,~\cite{renfaster,simonyanvery}). However, one drawback of DCNNs is that they rely on large collections of training data that have been annotated by humans, e.g., ImageNet~\cite{dengimagenet}. As such, DCNNs are restricted to domains for which large datasets of labeled examples are available. 

We explore the use of unsupervised sparse coding within a supervised DCNN to help alleviate the need for an abundance of training labels. Typically, the encodings (defined as the collection of activation values calculated from a layer) in DCNNs are computed via convolutions followed by a nonlinearity, and for a given task the weights are trained via supervised learning. In contrast, sparse coding~\cite{olshausensparse} aims to infer efficient, non-redundant encodings of a given input (e.g., photographs and videos), and the weights are trained via an unsupervised method. Inspired by theories of efficient coding in neural computation~\cite{barlowunsupervised}, sparse coding has been shown to exhibit similar properties to biological neurons in early stages of mammalian visual processing~\cite{olshausensparse,zhuvisual}. Modeling the accuracy and precision of biology in visual perception could provide novel insights into computer vision tasks, such as object detection. Additionally, the efficiency in representation exhibited by sparse coding can be advantageous in specialized, non-Von Neumann hardware architectures. Indeed, there has been work in implementations of sparse coding on low-power neuromorphic~\cite{knagsparse} and quantum~\cite{nguyensolving} hardware. 

One domain in which non-redundant encoding should be useful is in multi-view sensing. An efficient encoding should account for correlated offsets between different views of visual features. These offsets represent disparity in stereo images and optic flow in consecutive frames from a video. It follows that an encoding that accounts for such offsets should have some notion of depth~\cite{ferrismotion,qianbinocular}.

In this paper, we compare two types of convolutional network models that differ only in the first layer: (1) a \textit{sparse-coding} network, in which the weights and activations in the first layer are computed via unsupervised sparse coding, and (2) a \textit{supervised} network, in which the first-layer weights are learned via supervised training and activations are computed using these layer weights. In both network models, the weights and activations in all subsequent layers are computed via supervised convolutional layers. We show that sparse-coding networks is able to achieve better performance on a vehicle detection task on stereo-video data with a minimal amount of training labels. Additionally, we show that the performance of sparse-coding networks is more consistent---i.e., more robust to randomized order of training data and random initializations---than comparable supervised networks. Finally, we show that activations in the first (sparse-coding) layer in our sparse-coding networks are depth selective, which may provide an explanation for the differences in performance we observe in this study.

\subsection{Related Work} \label{sec:related}
Lundquist et al.~\cite{lundquistsparse} demonstrated that representations of stereo images obtained through sparse coding allow for an encoding that achieves better performance than a convolutional layer in the task of pixel-wise depth estimation. The authors show that the sparse encoding is inherently depth selective, whereas the convolutional encoding is not. In this paper, we extend this work to encode stereo-video clips and compare encodings on a vehicle-detection task.

A closely related study by Coates et al.~\cite{coatesanalysis} compared unsupervised and supervised methods with two-layer networks on an image classification task. The authors demonstrated that an unsupervised layer does not outperform a comparable layer with a convolutional encoding. While our experimental results agree with this finding in the case of two layers, we find that limiting the number of labeled training examples or adding an additional supervised layer in the sparse-coding network allows the network to outperform the supervised network.

Recent work by Lotter et al.~\cite{lotterdeep} shares the motivation of utilizing unsupervised learning to alleviate the need for labeled training data. Specifically, the authors use unsupervised learning to predict future frames of a video. They additionally show that their network achieves better performance than standard DCNNs when each is trained on only a limited amount of training data. In contrast to future frame prediction, our work aims to achieve image representation through sparse coding. Additionally, Lotter et al. uses a recurrent neural network~\cite{hochreiterlong} as the backbone of their network, whereas we use sparse coding as the unsupervised learning algorithm.

Other work~\cite{erhandoes,rainaself} explores the use of unsupervised learning techniques within a supervised network. However, most work in this area does not explore natural scenes (instead, focusing on datasets such as MNIST for handwritten digit recognition). Here, we extend this work to the domain of stereo video captured ``in the wild''. Additionally, we explicitly compare performance between the use of unsupervised learning versus supervised learning within a DCNN.

There has been other work in unsupervised learning of multi-view data~\cite{hoyerindependent,memisevicstereopsis,olshausensparsetime}. In contrast, our work aims to explicitly compare unsupervised learning to supervised learning for the task of vehicle detection in multi-view data.

\section{Sparse Coding}\label{sec:sparsecode}

Sparse coding aims to represent an input (e.g., an image) as a linear combination of basis vectors drawn from a provided set (referred to as a \textit{dictionary}) such that the original signal is recoverable with minimal degradation. Each basis vector is weighted by a scalar coefficient (referred to as an \textit{activation}), and the set of activations are taken to be the encoding of the input. Sparse coding constrains the activations to be sparse (i.e., to have few nonzero activations), such that resulting activations are non-redundant. Here, inferring the sparse set of activations is an optimization problem, unlike encodings calculated via supervised convolutional layers.

Formally, sparse coding aims to minimize the cost function
\begin{equation} \label{eq:sparsecode}
J(\boldsymbol{I}, \boldsymbol{a}, \boldsymbol{\Phi}) = 
\frac{1}{2} (\overbrace{\| \boldsymbol{I} - \boldsymbol{a} \circledast \boldsymbol{\Phi}\|_2}^{\text{Reconstruction error}})^2 + 
\lambda \overbrace{\| \boldsymbol{a} \|_1}^{\text{Sparsity term}}\text{.}
\end{equation}
Specifically, the algorithm aims to minimize the difference between a given input $\boldsymbol{I}$ and a reconstruction, where the difference is measured by Euclidean distance (i.e., $\|\cdot\|_2$, or the $L_2$ norm). The reconstruction $\boldsymbol{a} \circledast \boldsymbol{\Phi}$ is calculated via a linear combination of basis vectors drawn from a dictionary $\boldsymbol{\Phi}$, weighted by activation coefficients $\boldsymbol{a}$. Here, $\circledast$ denotes the transposed convolution operation~\cite{zeilerdeconvolutional}. The sparsity term constrains the activations $\boldsymbol{a}$ to be sparse, by measuring the sum of the absolute value of $\boldsymbol{a}$ (i.e., $\|\cdot\|_1$, or the $L_1$ norm)\footnote{The $L_1$ norm is used as a surrogate to the $L_0$ norm (i.e., the number of nonzero elements), as Equation~\ref{eq:sparsecode} is non-convex with respect to $\boldsymbol{a}$ if the $L_1$ norm is replaced with an $L_0$ norm.}. $\lambda$ is a hyperparameter that controls the trade-off between the reconstruction error and sparsity. 

The process of sparse coding that minimizes Equation~\ref{eq:sparsecode}, given a training set, is broken into two parts: (1) encoding an input by finding a set of activations $\boldsymbol{a}$ for input $\boldsymbol{I}$, and (2) learning a set of basis vectors (i.e., a dictionary) $\boldsymbol{\Phi}$ for the dataset. Encoding input $\boldsymbol{I}$ involves inferring activations $\boldsymbol{a}$ by minimizing Equation~\ref{eq:sparsecode} with respect to $\boldsymbol{a}$ while holding $\boldsymbol{\Phi}$ fixed, for which we use the biologically informed Locally Competitive Algorithm (LCA)~\cite{rozelllocally}. The final activations $\boldsymbol{a}$ are taken to be the output activations of the corresponding input $\boldsymbol{I}$. A sparse-coding layer can replace a convolutional layer in a DCNN by using activations $\boldsymbol{a}$ as the output of the layer. Learning a dictionary for sparse coding, analogous to learning filter weights in a DCNN, involves minimizing the cost function with respect to $\boldsymbol{\Phi}$ while holding $\boldsymbol{a}$ fixed via backpropagation of Equation~\ref{eq:sparsecode}. In the domain of images, dictionaries learned from sparse coding tend to represent oriented edges~\cite{olshausensparse}. In our method, the input is first encoded using LCA, which is followed by one gradient descent step of minimizing the cost function with respect to $\boldsymbol{\Phi}$ while holding $\boldsymbol{a}$ fixed. Basis vectors are normalized to have unit $L_2$ norm after each update. Updating $\boldsymbol{\Phi}$ is repeated for multiple input batches until convergence.

\section{Experiments} \label{sec:experiments}
We compare fully supervised networks with networks incorporating an unsupervised sparse-coding layer by testing performance on a vehicle detection task using stereo video. We test the effect of training set size by varying the number of labeled training examples available to the network. We used the KITTI object detection dataset~\cite{geigerare} for experiments. The dataset contains approximately 7000 examples, which we split into 6000 for training and 1000 for testing. Each example consists of three stereo frames ordered in time, with bounding box annotations for various objects in the left camera's last frame as ground truth. We normalize the stereo-video inputs to have zero mean and unit standard deviation and we downsampled them to be $256\times64$ pixels. We concatenated stereo inputs such that the input contains six features, i.e., RGB inputs from both left and right cameras. We kept time in a separate dimension for three-dimensional convolutions (for convolutional layers) or transposed convolution (for sparse-coding layers) across the time, height, and width axes of the input.

We generated the ground truth for our task by sliding a $32\times16$-pixel non-overlapping window across the left camera's last frame. We considered a window to be a positive instance if the window overlaps with any part of a car, van, or truck bounding box provided by the ground truth. The final output of each network is the probability of a window containing a vehicle, for all windows in the frame. We use the cross entropy between the ground truth and estimated probabilities as the supervised cost function to train all supervised layers within the network.

We tested various encoding schemes along with various weight initializations for the first layer of $n$-layer networks, as follows: 

\begin{itemize}
\item \mname{ConvSup}: Convolutional encoding. Weights are initialized randomly and learned via supervised training for car detection.
\item \mname{SparseUnsup}: Unsupervised sparse encoding to learn activations and weights.
\item \mname{ConvRand}: Convolutional encoding. Weights are initialized randomly and are not updated. This gives a random-weight baseline for the networks.
\item \mname{ConvUnsup}: Convolutional encoding. Weights are initialized from weights learned via unsupervised sparse coding and are not updated. This control tests the effect of weights versus encoding scheme on performance.
\item \mname{ConvFinetune}: Convolutional encoding. Weights are initialized from weights learned via unsupervised sparse coding. The weights were additionally trained via supervised training for car detection. This control is similar to ConvUnsup but tests the effect of additional training on the first-layer weights.
\end{itemize}

Once the first layer is set to one of the five possible options, the remaining $n-1$ layers contain convolutional layers learned via backpropagation of the supervised loss. Each model was repeated six times with different random initial conditions and random presentations of the training data to get a range in performance for each network.

All models, experiments, and hyperparameters are available online\footnote{\url{https://github.com/slundqui/TFSparseCode/}}.

\section{Results} \label{sec:results}
Table~\ref{tab:scores} shows the area under precision-versus-recall curve (AUC) of all models trained on all available training data, each tested with two, three, and four layers. Each network was trained on the car-detection task six times, with random weight initialization in higher layers and random ordering of training examples, in order to obtain the range of AUC values ("scores"). The range columns in Table 1 gives the difference between the maximum and minimum scores over the six runs. Here, we find that \mname{SparseUnsup} performs worse than \mname{ConvSup} and \mname{ConvFinetune} with two layers, which agrees with the findings of Coates et al.~\cite{coatesanalysis}. However, \mname{SparseUnsup} outperforms \mname{ConvSup} in networks with three or more layers. This difference in performance due to the number of layers is likely because of the lack in capacity of the two layer model to map input to detection.

One key finding is that \mname{SparseUnsup} is much more consistent (i.e., much less susceptible to random initial conditions and ordering of training examples) compared to all other models, as shown by the low range of AUC scores. Interestingly, \mname{SparseUnsup} has lower range in performance than \mname{ConvUnsup}, where both models use unsupervised weights learned via sparse coding and only differ in encoding scheme. This suggests that inferring activations in sparse coding is likely the reason for the additional consistency in performance.

\begin{figure}[!t] 
    \centering
    \includegraphics[width=.30\textwidth]{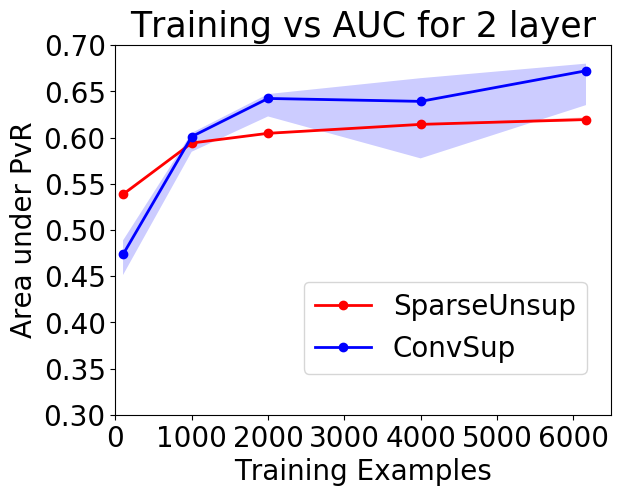}\hfill
    \includegraphics[width=.30\textwidth]{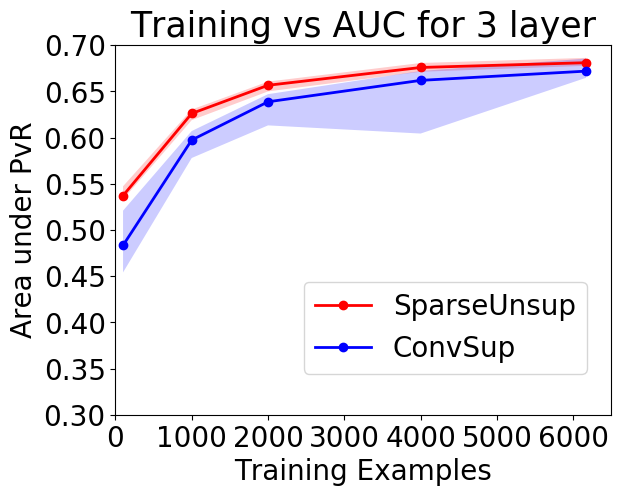}\hfill
    \includegraphics[width=.30\textwidth]{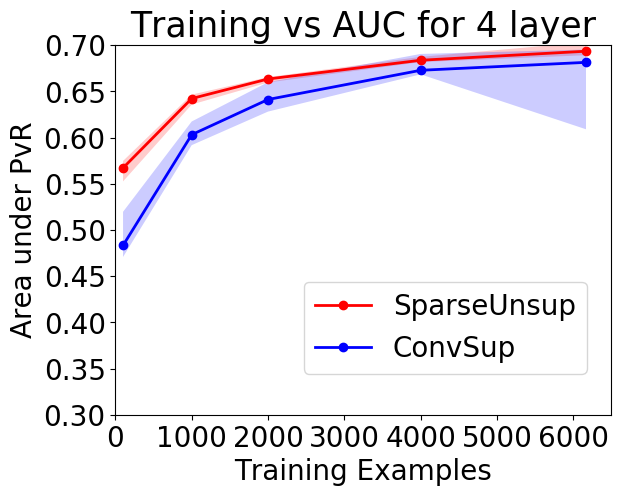}
    \caption[ ]{Number of training examples available versus AUC score for \mname{SparseUnsup} (red) and \mname{ConvSup} (blue) for two (left), three (middle), and four (right) layer networks. Each point is the median score over six independent runs, with the area between the maximum and minimum score filled in. Best viewed in color.}
    \label{fig:train_vs_auc_layers}
\end{figure}

Figure~\ref{fig:train_vs_auc_layers} gives the performance of \mname{SparseUnsup} and \mname{ConvSup} while varying the number of labeled training examples that each network was trained on. Here, the unsupervised weights were learned from all available data, without using training labels. We find that \mname{SparseUnsup} achieves better performance than \mname{ConvSup} across all numbers of labeled training examples tested for three and four layer networks. Additionally, \mname{SparseUnsup} achieves better performance with two layer networks when the number of training examples is limited to only 100 training examples. Finally, \mname{SparseUnsup} is much more consistent (as shown by the low range in AUC scores) in performance than that of \mname{ConvSup}, for all models tested.

\begin{figure}[!t]
    \centering
    \includegraphics[width=.95\textwidth]{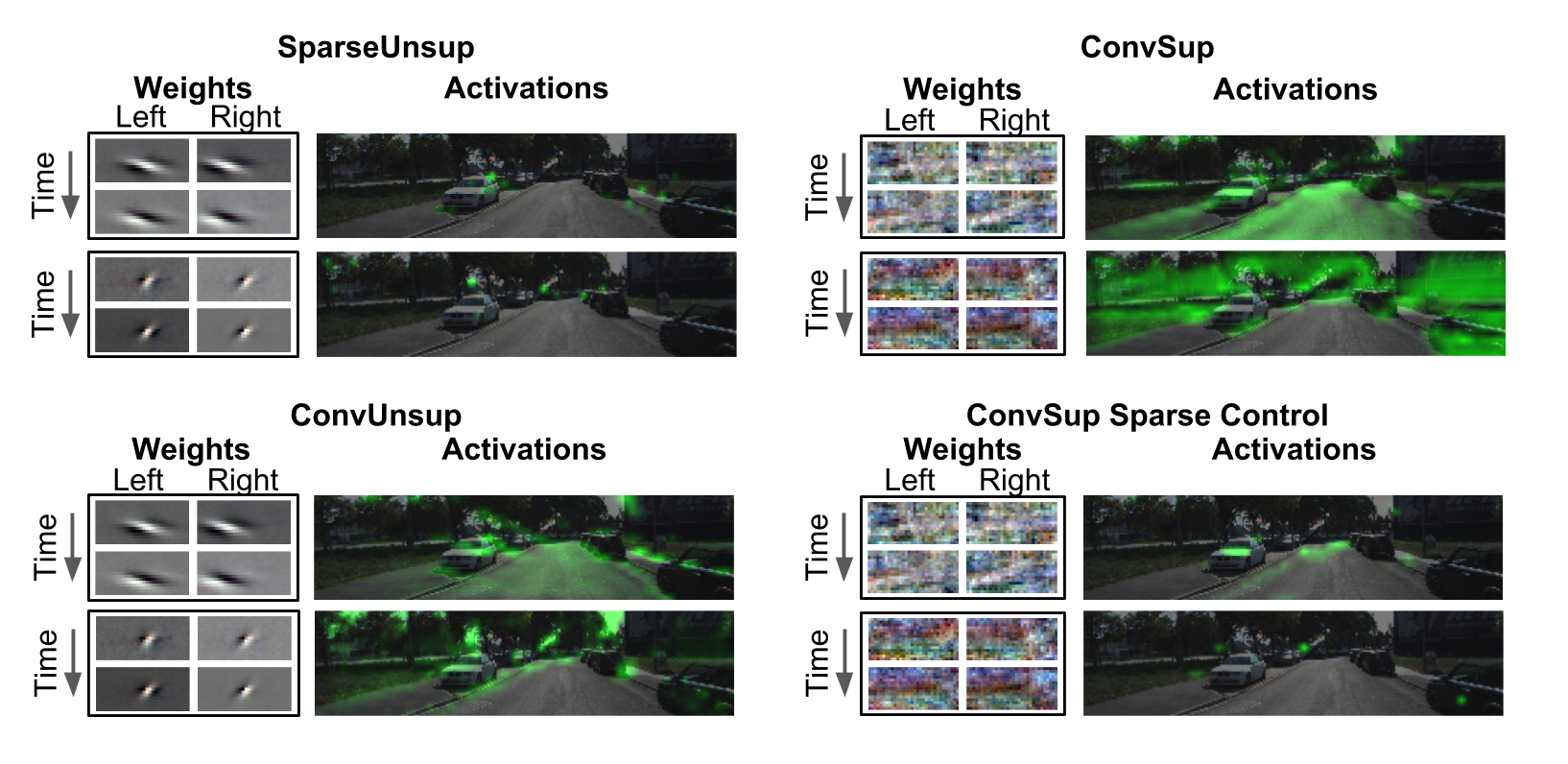}
    \caption[ ]{Nonzero activations of example weights overlaid on the input image. Magnitude of pixel values in green correspond to magnitude of activations. \mname{SparseUnsup}: Activations for near tuned (top) and far tuned (bottom) weights for the sparse-coding layer. \textbf{\mbox{ConvSup} Sparse Control}: Activations from \mname{ConvSup} with a threshold applied such that the number of activations matched that of sparse coding across the dataset. Best viewed in color.}
    \label{fig:featuremaps}
\end{figure}

We compare activation maps for the first layer of \mname{SparseUnsup}, \mname{ConvUnsup}, and \mname{ConvSup} in Figure~\ref{fig:featuremaps}. We find that the sparse-coding activations are selective to certain depths. For example, in Figure~\ref{fig:featuremaps} for \mname{SparseUnsup}, the top row shows a fast moving edge detector with a large binocular shift that corresponds to image features close to the camera, whereas the bottom row shows a static edge detector with no binocular shift that corresponds to image features far from the camera. In contrast, no convolutional layers show depth selectivity.

To control for the difference in number of nonzero activations, we applied a threshold to convolutional activations to match the number of nonzero activations in sparse coding, as shown in Figure~\ref{fig:featuremaps} \mname{ConvSup Sparse Control}. Sparse controls for \mname{ConvUnsup} produced no nonzero activations on this input. Here, we show that sparsity-controlled activations from \mname{ConvSup} do not show depth selectivity, as activations are active on image features at different depths. In all, these results may explain the differences in performance between convolutional layers and sparse-coding layers.

\begin{table}[!t]
\centering
\begin{tabular}{ l | c c | c c | c c }
Model                          & 2 layers   & Range        & 3 layers & Range       & 4 layers   & Range      \\ 
\hline
\mname{ConvSup}      & \bf{0.672} & 0.045        & 0.672    & 0.021       & 0.681      & 0.086       \\  
\mname{SparseUnsup}    & 0.619      & \bf{0.004}   & \bf{0.681}& \bf{0.009} & \bf{0.693} & \bf{0.014}  \\
\mname{ConvRand}     & 0.467      & 0.009        & 0.574    & 0.052       & 0.592      & 0.033       \\  
\mname{ConvUnsup}    & 0.561      & 0.044        & 0.609    & 0.028       & 0.609      & 0.033       \\  
\mname{ConvFinetune} & 0.660      & 0.020        & 0.641    & 0.081       & 0.691      & 0.117       \\  
\end{tabular}
\caption{Area under precision-versus-recall curve (AUC) for all models tested with varying depths. Models were ran six times with different initial conditions. Each value represents the median AUC score, and the range represents the difference between the highest score and the lowest score. \mname{Chance} scores at .221 AUC.} \label{tab:scores}
\end{table}

\section{Conclusion} \label{sec:discussion}
We have shown that a neural network that incorporates unsupervised learning is able to outperform a fully supervised network when there exists limited labeled training data. Additionally, we show that performance of fully supervised networks can vary substantially when compared to networks with a sparse-coding layer. Finally, we compare activations and show that depth selective activations emerge from applying sparse coding to stereo-video data. In all, these results show that unsupervised sparse coding can be useful in domains where there exists a limited amount of available labeled training data.

\subsubsection*{Acknowledgments} \label{sec:ack}
This research is funded by the DARPA Cooperative Agreement Award HR0011-13-2-0015. We would like to thank Dylan M. Paiton for valuable conversations pertaining to this manuscript.

\bibliographystyle{plain}
\bibliography{unsupervised}

\end{document}